\documentclass[11pt]{article}
\pdfoutput=1
\usepackage{authblk}
\usepackage{eacl2017}
\usepackage{times}
\usepackage{url}
\usepackage{latexsym}
\usepackage{amsmath}

\usepackage{scrextend} 
\usepackage{url}
\usepackage{color,floatrow}
\usepackage{booktabs}
\usepackage{lingmacros}
\usepackage{xspace}
\usepackage{graphicx}
\usepackage{adjustbox}
\usepackage[table]{xcolor}

\newcommand{\erstdt}{En-DT}
\newcommand{\enrst}{En-DT}
\newcommand{\pcc}{De-DT}
\newcommand{\derst}{De-DT}

\newcommand{\brrst}{Pt-DT}
\newcommand{\ptrst}{Pt-DT}
\newcommand{\barst}{Eu-DT}
\newcommand{\eurst}{Eu-DT}
\newcommand{\esrst}{Es-DT}
\newcommand{\sprst}{Es-DT}
\newcommand{\durst}{Nl-DT}
\newcommand{\nlrst}{Nl-DT}

\newcommand{\rel}[1]{\textit{#1}}
\newcommand{\crel}[1]{{\sc #1}\xspace}

\usepackage{fancyvrb}

\usepackage{tikz}
\usetikzlibrary{shapes.geometric}
\usetikzlibrary{positioning}
\usepackage{tikz-qtree,tikz-qtree-compat}
\tikzset{every tree node/.style={align=center, anchor=north}}
\usetikzlibrary{arrows}
\newfloatcommand{capbtabbox}{table}[][\FBwidth]

\eaclfinalcopy 

\title{Cross-lingual RST Discourse Parsing}
\author[1]{\bf Chlo{\'e} Braud}
\author[2]{\bf Maximin Coavoux}
\author[1]{\bf Anders S{\o}gaard}
\affil[1]{CoAStaL\\ DIKU\\University of Copenhagen\\ University Park 5, 2100 Copenhagen}
\affil[2]{LLF, CNRS\\ Univ Paris Diderot\\ Sorbonne Paris Cit\'{e}}
\affil[  ]{\tt  \{braud,soegaard\}@di.ku.dk}
\affil[ ]{\tt  \{maximin.coavoux\}@etu.univ-paris-diderot.fr}

\date{}

\begin{document}
\maketitle
\begin{abstract}
Discourse parsing is an integral part of understanding information flow and argumentative structure in documents. Most previous research has focused on inducing and evaluating models from the English RST Discourse Treebank.
However, discourse treebanks for other languages exist, including Spanish, German, Basque, Dutch and Brazilian Portuguese.
The treebanks share the same underlying linguistic theory, but differ slightly in the way documents are annotated.  
In this paper, we present (a) a new discourse parser which is simpler, yet competitive (significantly better on 2/3 metrics) to state of the art for English, (b) a harmonization of discourse treebanks across languages, enabling us to present (c) what to the best of our knowledge are the first experiments on cross-lingual discourse parsing. 

\end{abstract}

\section{Introduction}

\begin{figure}[!htb]
    \centering
		\begin{tikzpicture}[-,scale=1.2, auto,swap]
        \node (a) at (0,1) {1};
        \node (b) at (2,1) {2};
        \node (c) at (3,2) {3};
        \node (d) at (1,2) {NN-\crel{Comparison}};
        \node (e) at (2,3) {NS-\crel{Attribution}};
        \path (a) edge (d);
        \path (b) edge (d);
        \path (d) edge (e);
        \path (c) edge (e);
		\end{tikzpicture}
    \caption{Tree for the structure covering the segments $1$ to $3$ in document 1384 in the English RST Discourse Treebank.}
	\label{ex:fig}
\end{figure}
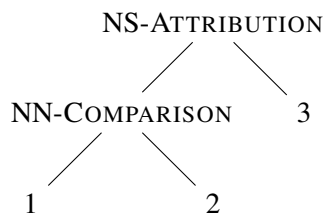

Documents can be analyzed as sequences of hierarchical discourse structures. Discourse structures describe the organization of documents in terms of discourse or rhetorical relations. 
For instance, the three discourse units below can be represented by the tree in Figure~\ref{ex:fig}, where a relation \crel{Comparison} holds between the segments 1 and 2, and a relation \crel{Attribution} links the segment covering the units 1 and 2, and the segment 3.\footnote{``NS" and ``NN" in Figure~\ref{ex:fig} describe the nuclearity of the segments, see Section~\ref{sec:rst}.} 
\begin{itemize}
\setlength\itemsep{.1em}
\item[1] Consumer spending in Britain rose 0.1\% in the third quarter from the second quarter
\item[2] and was up 3.8\% from a year ago, 
\item[3] the Central Statistical Office estimated.
\end{itemize}

Rhetorical Structure Theory (RST)~\cite{mann:rhetorical:1988} is a prominent linguistic theory of discourse structures, in which texts are analyzed as constituency trees, such as the one in Figure~\ref{ex:fig}. This theory guided the annotation of the RST Discourse Treebank (RST-DT)~\cite{carlson:building:2001} for English, from which several text-level discourse parsers have been induced~\cite{duverle:hilda:2010,joty:novel:2012,feng:linear:2014,li:recursive:2014,ji:representation:2014}. 
Such parsers have proven to be useful for various downstream applications~\cite{daume:noisy:2002,burstein:finding:2003,higgins:evaluationg:2004,thione:hybrid:2004,sporleder:discourse:2005,taboada:applications:2006,louis:discourse:2010,bhatia:better:2015}. 

There are discourse treebanks for other languages than English, including Spanish, German, Basque, Dutch, and Brazilian Portuguese. However, most research experimenting with these languages has focused on rule-based systems~\cite{pardo:development:2008,maziero:dizer:2011} or has been limited to intra-sentential relations~\cite{maziero:adaptation:2015}.

Moreover, all discourse corpora are limited in size, since annotation is complex and time consuming. 
This data sparsity makes learning hard, especially considering that discourse parsing involves several complex and interacting factors, ranging from syntax and semantics, to pragmatics.
We thus propose to harmonize existing corpora in order to leverage information by combining datasets in different languages.

\paragraph{Contributions} In this paper, we propose a new discourse parser that is significantly better than existing parsers for English on 2/3 standard metrics. Our parser relies on fewer features than previous work and is arguably algorithmically simpler. Moreover, we present the first end-to-end statistical discourse parsers for other languages than English (6 languages, in total). We also present the first experiments in cross-lingual discourse parsers, showing that discourse parsing is possible even when no or very little labeled data is available for the language of interest. We do so by harmonizing available discourse treebanks, enabling us to apply models across languages. 
We make the code and preprocessing scripts available for download at \url{https://bitbucket.org/chloebt/discourse}.

\section{Related Work}
\label{sec:related}

The first text-level discourse parsers were developed for English, relying mainly on hand-crafted rules and heuristics \cite{marcu:rhetorical:2000,carlson:building:2001}. 
\newcite[HILDA]{duverle:hilda:2010} greedily use SVM classifiers to make attachment and labeling decisions, building up a discourse tree.  \newcite[TSP]{joty:novel:2012} build a two-stage parsing system, training separate sequential models (CRF) for the intra- and the inter-sentential levels. These models jointly learn  the relation and the structure, and a CKY-like algorithm is used to find the optimal tree. \newcite{feng:linear:2014} use CRFs only as local models for the inter- and intra-sententials levels. 
For Brazilian Portuguese, for example, the first system, called DiZer~\cite{pardo:development:2008,maziero:dizer:2011}, was also rule-based, but there has been some work on using classification of intra-sentential relations \cite{maziero:adaptation:2015}. 

Recently studies have focused on building good representations of the data. \newcite{feng:textlevel:2012} introduced linguistic features, mostly syntactic and contextual ones. \newcite{li:recursive:2014}~used a recursive neural network that builds a representation for each clause based on the syntactic tree, and then apply two classifiers as in~\newcite{duverle:hilda:2010}. This leads to the best performing system for unlabeled structure ($85.0$ in F$_1$). The system presented by~\newcite[DPLP]{ji:representation:2014} jointly learns the representation and the task: a large margin classifier is used to learn the actions of a shift-reduce parser, optimizing at the same time the loss of the parser and a projection matrix that maps the bag-of-word representation of the discourse units into a new vector space. 
This system, however, only slightly outperforms the original bag-of-word representation. 
DPLP is the best performing discourse parser for labeled structure, $71.13$ in F$_1$ for nuclearity and $61.63$\% for relation.

Our system is similar to these last approaches in learning a representation using a neural network.  
However, we found that good performance can already be obtained without using all the words in the discourse units, resulting in a parser that is faster and easier to adapt, as demonstrated in our multilingual experiments,  see Section~\ref{sec:results}. 

\section{RST framework}
\label{sec:rst}

\paragraph{Discourse analysis}
In building a discourse structure, the text is first segmented into elementary discourse units (EDU), mostly clauses. EDUs are the smallest discourse units (DUs). Discourse relations are then used to build DUs, recursively. A non-elementary DU is called a complex discourse unit (CDU). The structure of a document is the set of linked DUs.
In this paper, we focus on the Rhetorical Structure Theory (RST), a theoretical framework proposed by~\newcite{mann:rhetorical:1988}.

\paragraph{Nuclearity}
A DU is either a \emph{nucleus} or a \emph{satellite}, the nucleus being the most important part of the relation (i.e. of the text), while the satellite contains additional, less important information. 
In general, this feature depends on the relation: a relation can be either mono-nuclear (with a scheme nucleus-satellite or satellite-nucleus depending on the relative order of the spans), or multi-nuclear.  
Some relations can be either mono- or multi-nuclear, such as \rel{consequence} or \rel{evaluation} in the RST-DT.

\begin{table*}[ht!]
\begin{tabular}{lrrrrrrrr}
\toprule
Corpus              & \#Doc         & \#Trees   & \#Words          & \#Rel & \#Lab  & \#EDU  & max/min/avg & \#CDU  \\
\midrule
\erstdt 			& 385           & {\bf 385}       & $206,300$         &56&  110 & 21,789 & 304/2/56.6   & 21,404  \\
\ptrst 				& 330 			& {\bf 329} 		& $135,820$	& 32 &  58	& 12,573 & 187/3/38.2 & 12,244 \\
\sprst\footnote{The test set contains $84$ documents doubly annotated, we report figures for annotator A. 
}
                    & 266           & {\bf 266}       & $69,787$          & 29    &  43       & 4,019 & 77/2/11.5     &  3,671  \\
\derst 				& 174           & {\bf 173}       & $32,274$          & 30    &  46       & 2,790 & 24/10/16.1    & 2,617  \\
\nlrst              & 80            & 80        & $27,920$          & 31    &  51       & 2,345 & 47/14/29.3    &  2,265  \\
\barst  		    & 88            & 85        & $27,982$          & 31    &  50       & 2,396 & 68/3/28.2     & 2,311   \\
\bottomrule
\end{tabular}
\caption{Number of documents (\#Doc), trees (\#Trees, less than \#Doc when we were unable to parse a document, see Section~\ref{subsec:harmonization}), words (\#Words, see Section~\ref{sec:setting}), relations (\#Rel, originally), labels (\#Lab, relation and nuclearity), EDUs (\#EDU, max/min/avg number of EDUs per document), and CDUs (\#CDU).
}
\label{table:stats}
\end{table*}

\paragraph{Binary trees}
In the original RST framework, each relation is associated with an application scheme that defines the nuclearity of the DUs (mono- or multi-nuclear relation), and the number of DUs linked.
Among the six schemes, two correspond to a link between more than two DUs, either a nucleus shared between two mono-nuclear relations (e.g. \rel{motivation} and \rel{enablement}) or a relation linking several nuclei (e.g. \rel{list}). 
\newcite{marcu:discourse:1997} proposed to simplify the representation to binary trees, and all discourse parsers are built on a binary representation.

\section{Data}
\label{sec:data}

We test our discourse parser on six languages, using available RST corpora  harmonized as described in Section~\ref{subsec:harmonization}.
Information about the datasets are summarized in Table~\ref{table:stats}.

\subsection{RST corpora}

\paragraph{English}

The RST~Discourse Treebank~\cite{carlson:discourse:2001}, from now on \erstdt, is the most widely used corpus to build discourse parsers.
It contains $385$ documents in English from the Wall Street Journal.
The relation set contains $56$ relations (ignoring nuclearity and embedding information\footnote{In this corpus, the embedded relations are annotated with a specific label (suffix ``-e") that we removed.}).
The inter-annotator agreement scores are $88.70$ for the unlabeled structure (score ``Span"), $77.72$ for the structure with nuclearity (``Nuclearity") and $65.75$ with relations (``Relation").\footnote{See Section~\ref{sec:setting} for a description of these metrics.}

\paragraph{Brazilian Portuguese}

We merged all the corpora annotated for Brazilian Portuguese, as in~\cite{maziero:adaptation:2015}, to form the \brrst.
The largest corpus is CST-News\footnote{\url{http://nilc.icmc.usp.br/CSTNews/login/?next=/CSTNews/}}~\cite{cardoso:cstnews:2011}, it is composed of $140$ documents from the news domain annotated with $31$ relations. 
Authors report agreement scores corresponding to nuclearity ($0.78$ in $F_1$) and relations ($0.66$).

The other corpora are: 
Summ-it\footnote{\url{http://www.inf.pucrs.br/ontolp/downloads-ontolpplugin.php}}~\cite{collovini:summit:2007} -- $50$ texts from science articles in a newspaper, annotated with $29$ relations; Rhetalho\footnote{\url{http://conteudo.icmc.usp.br/pessoas/taspardo/Projects.htm}\label{rhetalho}}~\cite{pardo:rhetalho:2005} -- $40$ texts from the computer science and news domains, annotated with $23$ relations; and CorpusTCC\footref{rhetalho}~\cite{pardo:construccao:2003,pardo:relaccoes:2004} -- $100$ introductions of scientific texts in computer science, annotated with $31$ relations.

\paragraph{Spanish}

The Spanish RST~DT\footnote{\url{http://corpus.iingen.unam.mx/rst/index_en.html}}~\cite{dacunha:spanish:2011}, from now on \sprst, contains $267$ texts written by specialists on different topics (e.g. astrophysics, economy, law, linguistics)
The relation set contains $29$ relations.
The authors report inter-annotator agreement of $86$\% in precision for the unlabeled structure, $82.46$\% for the structure with nuclearity and $76.81$\% with relations.

\paragraph{German}

The Postdam Commentary  Corpus 2.0\footnote{\url{http://angcl.ling.uni-potsdam.de/resources/pcc.html}}~\cite{stede:postdam:2004,stede:postdam:2014}, from now on \pcc, contains newspaper commentaries annotated at several levels.
A part of this corpus (MAZ) contains $175$ documents annotated within the  RST framework using $30$ relations.\footnote{We systematically ignore the first segment of each document, the title, that is not linked to the rest of the text.}

\paragraph{Dutch}
The corpus for Dutch~\cite{vliet:building:2011,redeker:multi:2012}, from now on \durst, contains $80$ documents from expository (encyclopedias and science news website) and persuasive (fund-raising letters and commercial advertisements) genres, annotated with $31$ relations.
The authors report an agreement of $0.83$ for discourse spans, $0.77$ for nuclearity and $0.70$ for relations.

\paragraph{Basque}

 The Basque RST~DT\footnote{\url{http://ixa2.si.ehu.es/diskurtsoa/en/}}~\cite{iruskieta:basque:2013}, from now on \barst, contains $88$ abstracts from three specialized domains -- medicine, terminology and science --, annotated with $31$ relations.
 The inter-annotator agreement is $81.67$\% for the identification of the CDU~\cite{iruskieta:qualitative:2015}, and $61.47$\% for the identification of the relations.

\paragraph{Other corpora} 
To the best of our knowledge, the only two non English corpora not included are the one annotated for Tamil~\cite{subalalitha:approach:2012} that we were unable to find, and the (intra-sentential) one developed for Chinese~\cite{wu:new:2016}, for which we were unable to produce RST trees since annotation does not contain nuclearity indications.

For English, there are corpora annotated for other domains than the one covered by the \erstdt.
We however leave out-of-domain evaluation for future work: it requires to decide how to use a corpus annotated only at the sentence level (SFU review corpus)\footnote{\url{https://www.sfu.ca/~mtaboada/research/SFU_Review_Corpus.html}}, or a corpus annotated with genre specific relations~\cite{subba:effective:2009}.

\subsection{Harmonization of the datasets}
\label{subsec:harmonization}

Recent discourse parsers built on the \erstdt\ are based on pre-processed data: the corpus contains only binary trees, with the large label set mapped to $18$ coarse-grained classes. 
In this section, we describe this pre-processing step for all corpora used. 
Discourse corpora have been released under three different file formats: \texttt{dis} (\erstdt), \texttt{lisp} (Rhetalho and CorpusTCC) and \texttt{rs3} (all remaining corpora). The first two ones are bracketed format, the third one is an XML encoding. 
In all cases, the trees encoded do not look like the one in Figure~\ref{ex:fig}: the relations are annotated on the daughter nodes, on the satellite for mono-nuclear relations, or on all the nuclei for multi-nuclear relations. Moreover, in the \texttt{rs3} format, the nuclearity of the segments is not directly annotated, it has to be retrieved using the type of the relation (indicated at the beginning of each file) and the previous principle.
Our pre-processing step leads to corpora with bracketed files representing directly the RST trees (as in Figure~\ref{ex:fig}) with stand-off annotation of the text of the EDUs.

Note that, even if harmonized, the corpora are not parallel, making it hard to use them to study language variations for the discourse level.
Some preliminary work exists on this question \cite{iruskieta:qualitative:2015}.

\paragraph{Pre-processing} Some documents (format \texttt{rs3}) contain several roots or empty segments. 
We were generally able to remove useless units, that is units that are not linked to other ones within the tree, except for one document in the CST corpus (two roots, both linked to other units). 

Another issue concerns unordered EDUs: the structure annotated contains nodes spanning non adjacent EDUs. 
In general, we were able to correct these cases, but we failed to automatically produce trees spanning only adjacent EDUs for three documents in the \barst, and one document in the \derst.

\paragraph{Binarization} All the corpora contain non-binary trees that we map to binary ones.
In the \erstdt, common cases of non-binarity are nodes whose daughters all hold  the same multi-nuclear relation  -- indicating that this relation spans multiple DUs, e.g. \rel{list}.\footnote{Recall that in the original format, the relation is not annotated on the parent node but on the children.}
In rare cases, the children are two satellites and a nucleus -- indicating that the nucleus is shared by the satellites. 
These configurations are the ones described in~\cite{marcu:discourse:1997} (see Section~\ref{sec:rst}), and choosing right or left-branching leads to a similar interpretation.
For the \erstdt, right-branching is the chosen strategy since~\cite{soricut:sentence:2003}.

\begin{figure}[!htb]
    \centering
		\begin{tikzpicture}[-,scale=1.2, auto,swap]
        \node (a) at (1,2) {$S_1$};
        \node (b) at (1,0) {$N_2$};
        \node (c) at (3,0) {$S_3$};
        \node (d) at (4,1) {$S_4$};
        \node (e) at (2,3) {SN-$R_i$};
        \node (f) at (2,1) {NS-$R_j$};
        \node (g) at (3,2) {NS-$R_k$};
        \path (a) edge (e);
        \path (g) edge (e);
        \path (b) edge (f);
        \path (c) edge (f);
        \path (f) edge (g);
        \path (d) edge (g);
		\end{tikzpicture}
		\caption{Binary tree for a node $X$ with $4$ children: $S_1$-$R_i$, $N_2$-Span, $S_3$-$R_j$ and $S_4$-$R_k$.}
		\label{fig:bin2}
\end{figure}
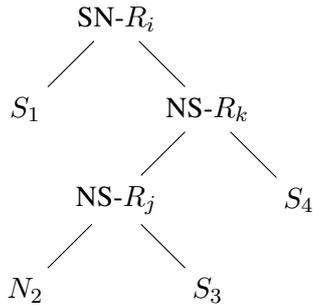

We found more diverse cases in the other corpora, and, for some of them, right-branching is impossible. 
It is the case when the daughters are one nucleus (annotated with ``Span", only indicating that this node spans several EDUs) and more than two satellites holding different relations -- i.e. the nucleus is shared by all the relations. 
More precisely, the issue arises when the last two children are satellites.
Using right-branching, we end with a node with two satellites as daughters, and thus a ill-formed tree.
In order to keep as often as possible the ``right-branching by default" strategy, we first do a right-branching and then a left-branching: beginning with four children -- $S_1$-$R_i$, $N_2$-Span, $S_3$-$R_j$ and $S_4$-$R_k$, indicating the relations $R_i(S_1,N_2)$, $R_j(N_2,S_3)$ and $R_k(N_2,S_4)$\footnote{$S$ being a satellite, $N$ a nucleus and $R$ a relation.}~--, we end up with the tree in Figure~\ref{fig:bin2}. 
Finally, we used a right-branching in all cases, except when the two last children are satellites.

\paragraph{Label set harmonization} We map all the relations used in the corpora to the $18$ coarse grained classes~\cite{carlson:discourse:2001} used to build the most recent discourse parsers on the \erstdt.\footnote{The full mapping is provided in Appendix~\ref{sec:supplemental}.} 

The mapping for the \erstdt\ is given in~\cite{carlson:discourse:2001}. 
For all the other corpora, we first map all the relations that exist in this mapping (i.e. used in the \erstdt\ annotation scheme) to their corresponding classes. 
We end with $18$ problematic relations, that is relations that were not used when annotating the \erstdt.

Among them, $10$ can be mapped easily, because they directly correspond to a class -- \rel{explanation} is mapped to the class \crel{Explanation}, \rel{elaboration} to \crel{Elaboration}, \rel{joint} to \crel{Joint} --, because they were just renamed -- \rel{reformulation} is mapped to the class \crel{Restatement} and \rel{solutionhood} (same as \rel{problem-solution}) to \crel{Topic-Comment} --, or because they correspond to a more-fine grained formulation of existing relations -- \rel{entity-elaboration} is mapped to \crel{Elaboration} and the $4$ \rel{volitional/non-volitional cause} and \rel{result} are mapped to the class \crel{Cause}, corresponding to the relations \rel{cause} and \rel{result} in the \erstdt.

For the remaining relations, we looked at the definition of the relations\footnote{\url{http://www.sfu.ca/rst/01intro/definitions.html}} to decide on a mapping.
Note that this label mapping is made quite easy by the fact that all the corpora were annotated following the same underlying theory -- they thus use relations defined using similar criteria --, and that we are using a coarse-grained classification -- we thus do not need to decide whether a relation is equivalent to another one, but rather whether it fits the properties of the other relations within a specific class.
Label mappings for corpora annotated following different frameworks are still discussed~\cite{roze:algebre:2013,benamara:mapping:2015}.  

We decided on the following mapping, considering the properties of the relations and the classes: \rel{parenthetical} -- used to give ``additional details" -- is mapped to \crel{Elaboration}, 
\rel{conjunction} -- similar to a \rel{list} with only two elements -- to \crel{Joint}, 
\rel{justify} -- similar to \rel{Explanation-argumentative} -- and \rel{motivation} -- quite similar to \rel{reason} and grouped with \rel{evidence} in~\cite{benamara:mapping:2015} -- to \crel{Explanation},  
\rel{preparation} -- presenting preliminary information, increasing the readiness to read the nucleus -- to \crel{Background}, 
and \rel{unconditional} and \rel{unless} -- linked to \rel{condition} -- to \crel{Condition}.

Finally, note that this mapping does not lead to having the same relation set for all the corpora, and that the relation distribution could vary among the datasets.

\section{Discourse Parser}
\label{sec:parser}

Our discourse parser builds discourse structures from segmented texts, we did not implement discourse segmenters for each language. Discourse segmenters only exist for English~\cite{duverle:hilda:2010} ($95,0$\% in F$_1$), Brazilian Portuguese~\cite{pardo:development:2008} ($56.8$\%) and Spanish~\cite{dacunha:diseg:2010,cunha:diseg:2012} ($80$\%). Discourse segmenters can be built quite easily relying only on manual rules as it is the case for the Spanish and Portuguese ones, especially considering that segmentation has generally been made coarser in the corpora built after the \enrst~\cite{vliet:building:2011}. While improving this first step is crucial, we focus on the harder step of tree building. 

\subsection{Description of the Parser}

We used the syntactic parser described in \newcite{coavoux-crabbe:2016:P16-1}, in the static oracle setting.
We chose this parser because it can take pre-trained embeddings as input and, more importantly,
because it was designed for morphologically rich languages and thus takes as input not only tokens and POS tags,
but any token attribute that is then mapped to a real-valued vector, which allows the use of complex features.

The parser is a transition-based constituent parser that uses a lexicalized shift-reduce transition system \cite{sagae-2005}.
The transition system is based on two data structures -- a \textit{stack} ($S$) stores partial trees and a \textit{queue} ($B$) contains the unparsed DUs.
A parsing \textit{configuration} is a couple $\langle S, B\rangle$.
In the initial configuration, $S$ is empty and $B$ contains the whole document.
The parser iteratively applies actions to the current configuration, in order
to derive new configurations until it reaches a final state, i.e.\ a parsing
configuration where $B$ is empty and $S$ contains a single element (the root of the tree).

The actions are defined as follows:
\begin{itemize}
    \item \textsc{Shift} pops an EDU from $B$ and pushes it onto $S$.
    \item \textsc{Reduce-R-X} and \textsc{Reduce-L-X} pop two DUs from $S$, push a new CDU with the label X on $S$ and assign its nucleus (Left or Right).
\end{itemize}

\paragraph{Scoring System} As in \newcite{chen-2014}, at each parsing step,
the parser scores actions with a feed-forward neural network.
The input of the network is a sequence of typed symbols extracted
from the top elements of $S$ and $B$.
The symbols are typically discourse relations or attributes of their nucleus EDU (e.g.\ first word of EDU, see Section \ref{sec:features}).

The first layer of the network projects these symbols onto an embedding
space (each type of symbol has its own embedding matrix).
The following two layers are non-linear layers with a ReLU activation.
The output of the network is a probability distribution over
possible actions computed by a softmax layer.

To generate a set of training examples $\{ a^{(i)}, c^{(i)}\}_{i=1}^N$,
we used the static oracle to extract the gold sequence of actions and configurations for each tree in the corpus.
The objective function of the parser is the negative
log-likelihood of gold actions given corresponding configurations:
\[ \mathcal{L(\boldsymbol \theta)} = - \sum_{i=1}^N \log P(a^{(i)} | c^{(i)}; \boldsymbol \theta) \]
where $\boldsymbol \theta$ is the set of all parameters, including embedding matrices.

We optimized this objective with the averaged stochastic gradient descent algorithm \cite{polyak-1992}.
At inference time, we used beam-search to find the best-scoring tree.

\subsection{Cross-lingual Discourse Parsing}
\label{subsec:cross}

Our first experiments are strictly monolingual, and they are intended to give state-of-the-art performance in a fully supervised setting.
We consider that we need at least $100$ documents to build a monolingual model, since we already keep around $65$ documents for test and development.

We then evaluate multi-source transfer methods, considering one language as the target and the others as sources.
More precisely, we will evaluate two settings: (1) training and optimizing only on the available source data; (2) training on all available data, including target ones if any, and optimizing on the development set of the target language. Setting (1) provides performance when no data are available at all in the target language, while (2) aims at evaluating if one can expect improvements by simply combining all the available data.

When combining the corpora, we cannot ignore lexical information as it has been done for syntactic parsing with delexicalized models \cite{mcdonald:multi:2011}.
Discourse parsing is a semantic task, at least when it comes to predict a rhetorical relation between two spans of text, and information from words have proven to be crucial \cite{rutherford:discovering:2014,braud:comparing:2015}.
We thus include word features using bilingual dictionaries -- i.e. translating the words used as features into a single language (English) --, or through cross-lingual word embeddings as proposed in \cite{guo:cross:2015} for dependency parsing. 
More precisely, we used the cross-lingual word representations presented in \cite{levy:strong:2017} that allow multi-source learning and have proven useful for POS tagging but also more semantic-oriented tasks, such as dependency parsing and document classification.

\subsection{Features}
\label{sec:features}

As in previous studies, we used features representing the two EDUs on the top of the stack and the EDU on the queue. If the stack contains CDUs, we use the nuclearity principle to choose the head EDU, converting multi-nuclear relations into nucleus-satellite ones as done since~\cite{sagae:analysis:2009}. 
However, we found that using these information also for the left and right children of the two CDUs on the top of the stack, and adding as a feature the representation built for these two CDUs lead to important improvements.

\paragraph{Lexical features}
We use the first three words and the last word along with their POS, features that have proven useful for discourse~\cite{pitler:automati:2009}, and the words in the \emph{head set}~\cite{sagae:analysis:2009} -- i.e. words whose head in the dependency graph is not in the EDU --, here limited to the first three.\footnote{Having more than three tokens in the head set is rare.} 
This head set contains the head of the sentence (in general, the main event), or words linked to the main clause when the segment does not contain the head (especially, discourse connectives that are subordinating or coordinating conjunctions could be found there).
The words are the boundaries could also contain discourse connectives, adverbs or temporal expressions that could be relevant for discourse structure.
Note however that these feature have been built for English, and they could be less useful for other languages.
We leave the question of investigating their utility linked to word order differences for future work.

Note that we do not use all the words in the EDUs as features, contrary to~\cite{li:recursive:2014,ji:representation:2014}. 
Our only word features are the words in the head set and at the boundaries, thus $7$ words per EDU.
When using word embeddings, we concatenate the vectors for each word, each of $d$ dimensions, keeping the same order to build a vector of $7d$ dimensions (e.g., the first word of the EDU corresponds to the first $d$ dimensions, the second has values between $d$ and $2d$). 

\paragraph{Position and length}
Other features are used to represent the position of the EDU in the document and its length in tokens.
We use thresholds to distinguish between very long (length $l>25$ tokens), long ($l>15$), short ($l>5$) and very short ($l\leq5$) EDUs. 
We also distinguish between the ``first" and the ``last" EDU in the document, and use also a threshold on the ratio $s=$(position of the EDU divided by the total number of EDUs) to separate EDUs at the beginning ($s < 0.25$), in the first middle ($0.25 \leq s < 0.5$), in the second middle ($0.5 \leq s < 0.75$) or in the end ($s>=0.75$). 

\paragraph{Position of the head}
We add a boolean feature indicating if the head of the sentence is in the current EDU or outside.

\paragraph{Number/date/percent/money}
We also use $4$ indicators of the presence of a date, a number, an amount of money and a percentage, features that have proven to be useful for discourse~\cite{pitler:automati:2009}. We build these features using simple regular expressions.

\section{Experiment settings}
\label{sec:setting}

\begin{table}
\begin{tabular}{lrrr}
\toprule
Corpus & Size dict. & \# words & \# unk. words \\
\midrule
\ptrst &18,049& 13,417 & 6,929\\
\esrst &22,815& 6,961 & 3,231\\
\derst &31,900& 5,856 & 1,762\\
\durst &19,012& 3,316 & 1,428\\
\eurst &1,092 & 6,553 & 5,446\\
\bottomrule
\end{tabular}
\caption{Dictionary coverage for each dataset on the train set when available, on the dev set else.}
\label{table:dictcoverage}
\end{table}

\paragraph{Data} For the \erstdt, we follow previous works in using the official test set of $38$ documents. For the \esrst, we report results on the test set A.\footnote{We found similar performance on the other test set.} For all the other corpora, we randomly choose $38$ documents to make a test set, and either use the remaining documents as development set (\nlrst\ and \eurst), or split them into a development set of $25$ documents, the remaining being used as training set (\erstdt, \esrst, \ptrst\ and \derst).

All the results given are based on a gold segmentation of the documents.

Each dataset is parsed using UDPipe,\footnote{\url{http://ufal.mff.cuni.cz/udpipe}} thus tokenizing, splitting into sentences and annotating each document based on the Universal Dependency scheme~\cite{ud13}. 

The word features for the non-English datasets are translated using available bilingual Wiktionaries\footnote{\url{https://en.wiktionary.org/wiki/User:Matthias_Buchmeier}} without disambiguation, the coverage of each dictionary is given in Table~\ref{table:dictcoverage}.
We also look for a translation of the lemma (and of the stems for the languages for which a stemmer\footnote{\url{https://pypi.python.org/pypi/snowballstemmer}} was available) as a backup strategy. 
When no translation is found, we keep the original token.

The word embeddings used were built on the EuroParl corpus~\cite{levy:strong:2017}. We keep only the $50$ first dimensions of the vectors representing the words, our preliminary experiments suggesting no significant differences against keeping the whole $200$ dimensions.
Unknown words are represented by the average vector of all word vectors.
For Basque, we had no access to these embeddings, we thus only report results using bilingual Wiktionaries.

\paragraph{Parameter tuning}
In our experiments we optimized on the development set the following parameters: the learning rate $\in \{0.01,0.02,0.03\}$, the learning rate decay constant $\in \{10^{-5},10^{-6},10^{-7},0\}$, the number of iterations $\in [1-20]$, and the size of the beam $\in \{1,2,4,8,16,32\}$. We fixed the number $N$ of hidden layers to $2$ and the size of the hidden layers $H$ to $128$ after experimenting on the \erstdt\ (with $N \in \{1,2,3\}$ and $H \in \{64,128\}$).

We fixed the size of the vectors for each feature to $50$ for word features,\footnote{When using embeddings, the final vector is of size $350$.} $16$ for POS, $6$ for position, $4$ for length, and $2$ for other features.

\paragraph{Metrics}

Following \cite{marcu:theory:2000} and most subsequent work, output trees are evaluated against gold trees in terms of how similar they bracket the EDUs (Span), how often they agree about nuclei when predicting a true bracket (Nuclearity), and in terms of the relation label, i.e., the overlap between the shared brackets between predicted and gold trees (Relation).\footnote{We use the evaluation script provided at \url{https://github.com/jiyfeng/DPLP}.} These scores are analogous to labeled and unlabeled syntactic parser evaluation metrics. 

\paragraph{Baseline}
Since we do not have state-of-the-art results for most of the languages, we provide results for a simple most frequent baseline (System MFS) that labels all nodes with the most frequent relation in the training or development set -- that is NN-\crel{Joint} for \derst\ and \esrst, and NS-\crel{Elaboration} for the others --, and build the structure by right-branching. 

\section{Results}
\label{sec:results}

\begin{table*}
\resizebox{\textwidth}{!}{
 	\begin{tabular}{l|rrr|rrr|rrr|rrr|rrr|rrr}
	\toprule
	System 	& \multicolumn{3}{c}{\enrst} & \multicolumn{3}{c}{\ptrst} & \multicolumn{3}{c}{\esrst} & \multicolumn{3}{c}{\derst} & \multicolumn{3}{c}{\nlrst} & \multicolumn{3}{c}{\eurst} \\
    		& Sp & Nuc & Rel & Sp & Nuc & Rel & Sp & Nuc & Rel & Sp & Nuc & Rel & Sp & Nuc & Rel & Sp & Nuc & Rel \\
	\midrule
	MFS 				& 58.2 & 33.4 & 22.1 & 57.3 & 33.9 & 23.23 & 82.0 & 51.5 & 17.7 & 61.3 & 37.8 & 13.2 & 57.9 & 35.5 & 22.0 & 63.2 & 34.9 & 18.8 \\
    \arrayrulecolor{black!30}\midrule
   Li et al.\footnote{\label{first}Scores reported from~\cite{li:recursive:2014}, and DPLP~\cite{ji:representation:2014}.} 			& 85.0  & 70.8  & 58.6 &- &- &- &- &- &- &- &- &- &- &- &- &- &- &- \\
    DPLP\footref{first} 	& 82.1 & 71.1 & {\bf 61.6} &- &- &- &- &- &- &- &- &- &- &- &- &- &- &- \\
    \arrayrulecolor{black!30}\midrule
    
    Mono 				& 85.0 & 72.3 & 60.1 & {\bf 82.0} & {\bf 65.1} & {\bf 49.9} & {\bf 89.7} & {\bf 72.7} & {\bf 54.4} & {\bf 80.2} & {\bf 53.9} & {\bf 35.0} & -&- &- &- &- &- \\
    + emb. 			& 83.5 & 68.5 & 55.9 & 81.3	& 62.9 & 48.8 & 89.3 & 72.4 & 51.4 & 77.7 & 51.6 & 31.1 &- &- &- &- &- &- \\
    \arrayrulecolor{black!30}\midrule
    Cross 		& 76.3 & 50.5 & 31.3 & 76.5 & 54.6 & 35.5 & 78.1 & 45.4 & 27.0 & 76.0 & 46.0 & 26.1 & 69.5 & 42.1 & 25.3 & {\bf 78.6} & {\bf 53.0} & 26.4 \\
    + dev. 		& \textbf{85.1} & \textbf{73.1} & 61.4 & 81.9 & 65.1 & 49.8 & 88.8 & 68.0 & 50.4 & 79.6 & 53.6 & 34.1 & {\bf 69.2} & {\bf 43.4} & {\bf 28.3} & 76.7 & 50.5 & {\bf 29.5} \\
    
    \arrayrulecolor{black!30}\midrule
    Human\footnote{For Brazilian Portuguese, inter-annotator agreement scores are only available for the CST-news corpus ; For Spanish, only precision scores are reported ; For Basque, the scores reported are different~\cite{iruskieta:qualitative:2015}.}		
    					& 88.7 & 77.7 & 65.8 & - & 78 & 66 & 86 & 82.5 & 76.8 & -&- &- & 83 &77 & 70 &81.7 & -&61.5 \\
	\arrayrulecolor{black}\bottomrule
	\end{tabular}}%
    \caption{Performance of our monolingual and cross-lingual systems for Span (Sp), Nuclearity (Nuc) and Relation (Rel). ``MFS" corresponds to the baseline system described in Section~\ref{sec:setting}; ``+~emb." is the monolingual system using word embeddings; ``+dev." means that the system is optimized on the development set of the target language (\textit{vs} the union of the source development sets). For cross-lingual systems, we only report our best results using either word embeddings or bilingual dictionaries.}
    \label{tab:results}
\end{table*}

\paragraph{Monolingual experiments}
Monolingual experiments are aimed at evaluating performance for languages having a large annotated corpus (at least $100$ documents).
Our results are summarized in Table~\ref{tab:results}. 
Our parser is competitive with state-of-the-art systems for English (first line in Table~\ref{tab:results}), with even better performance for unlabeled structure ($85.04$\%) and structure labeled with nuclearity ($72.29$\%).
These results show that using all the words in the units~\cite{ji:representation:2014,li:recursive:2014}, is not as useful as using more contextual information, that is taking more DUs into account (left and right children of the CDUs in the stack). 
However, the slight drop for Relation shows that we probably miss some lexical information, or that we need to choose a more effective combination scheme than concatenation.
We plan to use bi-LSTM encoders \cite{Hochreiter:Schmidhuber:97} to construct fixed-length representations of EDUs.

For the other languages, performance are still high for unlabeled structure, but far lower for labeled structure except for Spanish.
For this language, the quite high performance obtained were unexpected, since the corpus is far much smaller than the Portuguese one.
One possible explanation is that the Portuguese corpus is in fact a mix of different corpora, with varied domains, and possibly changes in annotation choices. 
On the other hand, the low results for German show the sparsity issue since it is the language for which we have the fewest annotations (``\#CDU", see Table~\ref{table:stats}).

\paragraph{Cross-lingual experiments}
When only relying on data from different languages (``Cross" in Table~\ref{tab:results}), we observe a large drop in performance compared to monolingual systems.
The source-only discourse parsers still have fairly high performance for unlabeled structure (around $70$\% or higher), the scores being especially low for relation identification.
This could indicate that our representation does not generalize well.
But it also comes from differences among the corpora. For example, only the \enrst\ and the \ptrst\ use the relation \crel{Attribution}.
This leads to a large drop in performance associated with this relation, when one of these corpora is not in the training data, especially for the source-only system for the \enrst\ (from $93$\% in $F_1$ to $30$\%).
On the other hand, on the \enrst, we observe improvement for other relations either largely represented in all the corpora (e.g. \crel{Joint} $+3$\%), or under-represented in the \enrst\ (e.g. \crel{Condition} $+3$\%).

When combining corpora for source and target languages (``+ dev." in Table~\ref{tab:results}), we obtain our best performing system for English, with all scores improved compared to our best monolingual system ($+0.8$ for Nuclearity and $+1.3$ for Relation). Otherwise scores are similar to the monolingual case. 

Finally, for languages without training set (\nlrst\ and \eurst), this strategy allows us to build parsers outperforming our simple baseline (MFS) by around $11-13$\% for Span, $8-15$\% for Nuclearity and $6-11$\% for Relation.
Having at least some annotated data to make a development set allows improvements against only using corpora in other languages (around $+3$\% for the \nlrst\ and the \eurst\ for Relation). On the other hand, we probably overfit our development data for the \eurst, since better results were obtained for unlabeled structure ($+2$\%) and structure with nuclearity ($+2.5$\%) using only data in other languages.

\paragraph{Word embeddings}
Using word embeddings (``+emb" in Table~\ref{tab:results}) for monolingual systems often leads to an important drop in performance, especially for Relation (from $-1.1$ to $-4.2$\%).
This demonstrates that these embeddings do not provide the large range of information needed for relation identification, a task inherently semantic.
We believe however that the results are not too low to prevent for interesting applications. 
It is noteworthy that the English parser with embeddings is still better than the systems proposed in~\cite{duverle:hilda:2010,joty:combining:2013}.

For cross-lingual experiments, the bilingual dictionaries perform generally better than embeddings (except for \ptrst\ and \derst\ for source-only systems), demonstrating again that we need representations more tailored to the task to leverage all relevant lexical information. 

\section{Conclusion}

We introduced a new discourse parser that obtains state-of-the-art performance for English. We harmonized discourse treebanks for several languages, enabling us to present results for five other languages for which available corpora are smaller, including the first cross-lingual discourse parsing results in the literature.

\section*{Acknowledgements}

We thank the three anonymous reviewers for their comments.  Chlo{\'e}
Braud and Anders S{\o}gaard were funded by the ERC Starting Grant LOWLANDS No. 313695.

\bibliography{discourse}

\begin{thebibliography}{}

\bibitem[\protect\citename{Benamara and Taboada}2015]{benamara:mapping:2015}
Farah Benamara and Maite Taboada.
\newblock 2015.
\newblock Mapping different rhetorical relation annotations: A proposal.
\newblock In {\em Proceedings of Starsem}.

\bibitem[\protect\citename{Bhatia \bgroup et al.\egroup
  }2015]{bhatia:better:2015}
Parminder Bhatia, Yangfeng Ji, and Jacob Eisenstein.
\newblock 2015.
\newblock Better document-level sentiment analysis from {RST} discourse
  parsing.
\newblock In {\em Proceedings of EMNLP}.

\bibitem[\protect\citename{Braud and Denis}2015]{braud:comparing:2015}
Chlo\'{e} Braud and Pascal Denis.
\newblock 2015.
\newblock Comparing word representations for implicit discourse relation
  classification.
\newblock In {\em Proceedings of EMNLP}.

\bibitem[\protect\citename{Burstein \bgroup et al.\egroup
  }2003]{burstein:finding:2003}
Jill Burstein, Daniel Marcu, and Kevin Knight.
\newblock 2003.
\newblock Finding the write stuff: automatic identification of discourse
  structure in student essays.
\newblock {\em IEEE Intelligent Systems: Special Issue on Advances in Natural
  Language Processing}, 18.

\bibitem[\protect\citename{Cardoso \bgroup et al.\egroup
  }2011]{cardoso:cstnews:2011}
Paula~C.F. Cardoso, Erick~G. Maziero, María Lucía~Castro Jorge, Eloize~R.M.
  Seno, Ariani Di~Felippo, Lucia Helena~Machado Rino, Maria das Gracas~Volpe
  Nunes, and Thiago A.~S. Pardo.
\newblock 2011.
\newblock {CSTN}ews - a discourse-annotated corpus for single and
  multi-document summarization of news texts in {B}razilian {P}ortuguese.
\newblock In {\em Proceedings of the 3rd RST Brazilian Meeting}, pages 88--105.

\bibitem[\protect\citename{Carlson and Marcu}2001]{carlson:discourse:2001}
Lynn Carlson and Daniel Marcu.
\newblock 2001.
\newblock Discourse tagging reference manual.
\newblock Technical report, University of Southern California Information
  Sciences Institute.

\bibitem[\protect\citename{Carlson \bgroup et al.\egroup
  }2001]{carlson:building:2001}
Lynn Carlson, Daniel Marcu, and Mary~Ellen Okurowski.
\newblock 2001.
\newblock Building a discourse-tagged corpus in the framework of {R}hetorical
  {S}tructure {T}heory.
\newblock In {\em Proceedings of the Second {SIGdial} Workshop on Discourse and
  Dialogue}.

\bibitem[\protect\citename{Chen and Manning}2014]{chen-2014}
Danqi Chen and Christopher~D Manning.
\newblock 2014.
\newblock A fast and accurate dependency parser using neural networks.
\newblock In {\em Empirical Methods in Natural Language Processing (EMNLP)}.

\bibitem[\protect\citename{Coavoux and
  Crabb\'{e}}2016]{coavoux-crabbe:2016:P16-1}
Maximin Coavoux and Benoit Crabb\'{e}.
\newblock 2016.
\newblock Neural greedy constituent parsing with dynamic oracles.
\newblock In {\em Proceedings of the 54th Annual Meeting of the Association for
  Computational Linguistics (Volume 1: Long Papers)}, pages 172--182, Berlin,
  Germany, August. Association for Computational Linguistics.

\bibitem[\protect\citename{Collovini \bgroup et al.\egroup
  }2007]{collovini:summit:2007}
Sandra Collovini, Thiago~I Carbonel, Juliana~Thiesen Fuchs, Jorge~C{\'e}sar
  Coelho, L{\'u}cia Rino, and Renata Vieira.
\newblock 2007.
\newblock Summ-it: Um corpus anotado com informa{\c{c}}oes discursivas visandoa
  sumariza{\c{c}}ao autom{\'a}tica.
\newblock {\em Proceedings of TIL}.

\bibitem[\protect\citename{da Cunha \bgroup et al.\egroup
  }2010]{dacunha:diseg:2010}
Iria da~Cunha, Eric SanJuan, Juan-Manuel Torres-Moreno, Marina Lloberas, and
  Irene Castell{\'o}n.
\newblock 2010.
\newblock Di{S}eg: Un segmentador discursivo autom{\'a}tico para el
  espa{\~n}ol.
\newblock {\em Procesamiento del lenguaje natural}, 45:145--152.

\bibitem[\protect\citename{da Cunha \bgroup et al.\egroup
  }2011]{dacunha:spanish:2011}
Iria da~Cunha, Juan{-}Manuel Torres{-}Moreno, and Gerardo Sierra.
\newblock 2011.
\newblock On the development of the {RST} {S}panish {T}reebank.
\newblock In {\em Proceedings of the Fifth Linguistic Annotation Workshop,
  {LAW}}.

\bibitem[\protect\citename{da Cunha \bgroup et al.\egroup
  }2012]{cunha:diseg:2012}
Iria da~Cunha, Eric SanJuan, Juan{-}Manuel Torres{-}Moreno, Marina Lloberes,
  and Irene Castell{\'{o}}n.
\newblock 2012.
\newblock Di{S}eg 1.0: The first system for {S}panish discourse segmentation.
\newblock {\em Expert Syst. Appl.}, 39(2):1671--1678.

\bibitem[\protect\citename{Daum{\'{e}}~III and Marcu}2009]{daume:noisy:2002}
Hal Daum{\'{e}}~III and Daniel Marcu.
\newblock 2009.
\newblock A noisy-channel model for document compression.
\newblock In {\em Proceedings of ACL}.

\bibitem[\protect\citename{Feng and Hirst}2012]{feng:textlevel:2012}
Vanessa~Wei Feng and Graeme Hirst.
\newblock 2012.
\newblock Text-level discourse parsing with rich linguistic features.
\newblock In {\em Proceedings of ACL}.

\bibitem[\protect\citename{Feng and Hirst}2014]{feng:linear:2014}
Vanessa~Wei Feng and Graeme Hirst.
\newblock 2014.
\newblock A linear-time bottom-up discourse parser with constraints and
  post-editing.
\newblock In {\em Proceedings of ACL}.

\bibitem[\protect\citename{Guo \bgroup et al.\egroup }2015]{guo:cross:2015}
Jiang Guo, Wanxiang Che, David Yarowsky, Haifeng Wang, and Ting Liu.
\newblock 2015.
\newblock Cross-lingual dependency parsing based on distributed
  representations.
\newblock In {\em Proceedings of ACL-IJCNLP}.

\bibitem[\protect\citename{Hernault \bgroup et al.\egroup
  }2010]{duverle:hilda:2010}
Hugo Hernault, Helmut Prendinger, David~A. duVerle, and Mitsuru Ishizuka.
\newblock 2010.
\newblock {HILDA}: A discourse parser using support vector machine
  classification.
\newblock {\em Dialogue and Discourse}, 1:1--33.

\bibitem[\protect\citename{Higgins \bgroup et al.\egroup
  }2004]{higgins:evaluationg:2004}
Derrick Higgins, Jill Burstein, Daniel Marcu, and Claudia Gentile.
\newblock 2004.
\newblock Evaluating multiple aspects of coherence in student essays.
\newblock In {\em Proceedings of HLT-NAACL}.

\bibitem[\protect\citename{Hochreiter and
  Schmidhuber}1997]{Hochreiter:Schmidhuber:97}
Sepp Hochreiter and J\"{u}rgen Schmidhuber.
\newblock 1997.
\newblock Long short-term memory.
\newblock {\em Neural Computation}, 9(8):1735--1780.

\bibitem[\protect\citename{Iruskieta \bgroup et al.\egroup
  }2013]{iruskieta:basque:2013}
Mikel Iruskieta, Mar\'{i}a~J. Aranzabe, Arantza Diaz~de Ilarraza, Itziar
  Gonzalez-Dios, Mikel Lersundi, and Oier Lopez de~la Calle.
\newblock 2013.
\newblock The {RST} {B}asque {T}reebank: an online search interface to check
  rhetorical relations.
\newblock In {\em Proceedings of the 4th Workshop RST and Discourse Studies}.

\bibitem[\protect\citename{Iruskieta \bgroup et al.\egroup
  }2015]{iruskieta:qualitative:2015}
Mikel Iruskieta, Iria da~Cunha, and Maite Taboada.
\newblock 2015.
\newblock A qualitative comparison method for rhetorical structures:
  identifying different discourse structures in multilingual corpora.
\newblock In {\em Proceedings of LREC}.

\bibitem[\protect\citename{Ji and Eisenstein}2014]{ji:representation:2014}
Yangfeng Ji and Jacob Eisenstein.
\newblock 2014.
\newblock Representation learning for text-level discourse parsing.
\newblock In {\em Proceedings of ACL}.

\bibitem[\protect\citename{Joty \bgroup et al.\egroup }2012]{joty:novel:2012}
Shafiq~R. Joty, Giuseppe Carenini, and Raymond~T. Ng.
\newblock 2012.
\newblock A novel discriminative framework for sentence-level discourse
  analysis.
\newblock In {\em Proceedings of EMNLP}.

\bibitem[\protect\citename{Joty \bgroup et al.\egroup
  }2013]{joty:combining:2013}
Shafiq~R. Joty, Giuseppe Carenini, Raymond~T. Ng, and Yashar Mehdad.
\newblock 2013.
\newblock Combining intra- and multi-sentential rhetorical parsing for
  document-level discourse analysis.
\newblock In {\em Proceedings of ACL}.

\bibitem[\protect\citename{Levy \bgroup et al.\egroup }2017]{levy:strong:2017}
Omer Levy, Anders S{\o}gaard, and Yoav Goldberg.
\newblock 2017.
\newblock A strong baseline for learning cross-lingual word embeddings from
  sentence alignments.
\newblock In {\em Proceedings of EACL}.

\bibitem[\protect\citename{Li \bgroup et al.\egroup }2014]{li:recursive:2014}
Jiwei Li, Rumeng Li, and Eduard~H. Hovy.
\newblock 2014.
\newblock Recursive deep models for discourse parsing.
\newblock In {\em Proceedings of EMNLP}.

\bibitem[\protect\citename{Louis \bgroup et al.\egroup
  }2010]{louis:discourse:2010}
Annie Louis, Aravind Joshi, and Ani Nenkova.
\newblock 2010.
\newblock Discourse indicators for content selection in summarization.
\newblock In {\em Proceedings of SIGDIAL}.

\bibitem[\protect\citename{Mann and Thompson}1988]{mann:rhetorical:1988}
William~C. Mann and Sandra~A. Thompson.
\newblock 1988.
\newblock {R}hetorical {S}tructure {T}heory: Toward a functional theory of text
  organization.
\newblock {\em Text}, 8:243--281.

\bibitem[\protect\citename{Marcu}1997]{marcu:discourse:1997}
Daniel Marcu.
\newblock 1997.
\newblock From discourse structures to text summaries.
\newblock In {\em Proceedings of the ACL Workshop on Intelligent Scalable Text
  Summarization}, pages 82--88.

\bibitem[\protect\citename{Marcu}2000a]{marcu:rhetorical:2000}
Daniel Marcu.
\newblock 2000a.
\newblock The rhetorical parsing of unrestricted texts: A surface-based
  approach.
\newblock {\em Computational Linguistics}.

\bibitem[\protect\citename{Marcu}2000b]{marcu:theory:2000}
Daniel Marcu.
\newblock 2000b.
\newblock {\em The Theory and Practice of Discourse Parsing and Summarization}.
\newblock MIT Press, Cambridge, MA, USA.

\bibitem[\protect\citename{Maziero \bgroup et al.\egroup
  }2011]{maziero:dizer:2011}
Erick~G. Maziero, Thiago A.~S. Pardo, Iria da~Cunha, Juan-Manuel Torres-Moreno,
  and Eric SanJuan.
\newblock 2011.
\newblock Di{Z}er 2.0-an adaptable on-line discourse parser.
\newblock In {\em Proceedings of 3rd RST Brazilian Meeting}, pages 1--17.

\bibitem[\protect\citename{Maziero \bgroup et al.\egroup
  }2015]{maziero:adaptation:2015}
Erick~G. Maziero, Graeme Hirst, and Thiago A.~S. Pardo.
\newblock 2015.
\newblock Adaptation of discourse parsing models for {P}ortuguese language.
\newblock In {\em Proceedings of the Brazilian Conference on Intelligent
  Systems (BRACIS)}.

\bibitem[\protect\citename{McDonald \bgroup et al.\egroup
  }2011]{mcdonald:multi:2011}
Ryan McDonald, Slav Petrov, and Keith Hall.
\newblock 2011.
\newblock Multi-source transfer of delexicalized dependency parsers.
\newblock In {\em Proceedings of EMNLP}.

\bibitem[\protect\citename{Nivre \bgroup et al.\egroup }2016]{ud13}
Joakim Nivre, {\v Z}eljko Agi{\'c}, Lars Ahrenberg, Maria~Jesus Aranzabe,
  Masayuki Asahara, Aitziber Atutxa, Miguel Ballesteros, John Bauer, Kepa
  Bengoetxea, Yevgeni Berzak, Riyaz~Ahmad Bhat, Cristina Bosco, Gosse Bouma,
  Sam Bowman, G{\"u}l{\c s}en Cebiroğlu~Eryiğit, Giuseppe G.~A. Celano, {\c
  C}ağrı {\c C}{\"o}ltekin, Miriam Connor, Marie-Catherine de~Marneffe,
  Arantza Diaz~de Ilarraza, Kaja Dobrovoljc, Timothy Dozat, Kira Droganova,
  Toma{\v z} Erjavec, Rich{\'a}rd Farkas, Jennifer Foster, Daniel Galbraith,
  Sebastian Garza, Filip Ginter, Iakes Goenaga, Koldo Gojenola, Memduh
  Gokirmak, Yoav Goldberg, Xavier G{\'o}mez~Guinovart, Berta
  Gonz{\'a}les~Saavedra, Normunds Gr{\= u}z{\=i}tis, Bruno Guillaume, Jan
  Haji{\v c}, Dag Haug, Barbora Hladk{\'a}, Radu Ion, Elena Irimia, Anders
  Johannsen, H{\"u}ner Ka{\c s}ıkara, Hiroshi Kanayama, Jenna Kanerva, Boris
  Katz, Jessica Kenney, Simon Krek, Veronika Laippala, Lucia Lam, Alessandro
  Lenci, Nikola Ljube{\v s}i{\'c}, Olga Lyashevskaya, Teresa Lynn, Aibek
  Makazhanov, Christopher Manning, C{\u a}t{\u a}lina M{\u a}r{\u a}nduc, David
  Mare{\v c}ek, H{\'e}ctor Mart{\'i}nez~Alonso, Jan Ma{\v s}ek, Yuji Matsumoto,
  Ryan {McDonald}, Anna Missil{\"a}, Verginica Mititelu, Yusuke Miyao,
  Simonetta Montemagni, Keiko~Sophie Mori, Shunsuke Mori, Kadri Muischnek, Nina
  Mustafina, Kaili M{\"u}{\"u}risep, Vitaly Nikolaev, Hanna Nurmi, Petya
  Osenova, Lilja {\O}vrelid, Elena Pascual, Marco Passarotti, Cenel-Augusto
  Perez, Slav Petrov, Jussi Piitulainen, Barbara Plank, Martin Popel, Lauma
  Pretkalniņa, Prokopis Prokopidis, Tiina Puolakainen, Sampo Pyysalo,
  Loganathan Ramasamy, Laura Rituma, Rudolf Rosa, Shadi Saleh, Baiba
  Saul{\=i}te, Sebastian Schuster, Wolfgang Seeker, Mojgan Seraji, Lena
  Shakurova, Mo~Shen, Natalia Silveira, Maria Simi, Radu Simionescu, Katalin
  Simk{\'o}, Kiril Simov, Aaron Smith, Carolyn Spadine, Alane Suhr, Umut
  Sulubacak, Zsolt Sz{\'a}nt{\'o}, Takaaki Tanaka, Reut Tsarfaty, Francis
  Tyers, Sumire Uematsu, Larraitz Uria, Gertjan van Noord, Viktor Varga,
  Veronika Vincze, Jing~Xian Wang, Jonathan~North Washington, Zden{\v e}k {\v
  Z}abokrtsk{\'y}, Daniel Zeman, and Hanzhi Zhu.
\newblock 2016.
\newblock Universal dependencies 1.3.
\newblock {LINDAT}/{CLARIN} digital library at Institute of Formal and Applied
  Linguistics, Charles University in Prague.

\bibitem[\protect\citename{Pardo and Nunes}2003]{pardo:construccao:2003}
Thiago A.~S. Pardo and Maria das Gra{\c{c}}as~Volpe Nunes.
\newblock 2003.
\newblock A constru{\c{c}}{\~a}o de um corpus de textos cient{\'\i}ficos em
  {P}ortugu{\^e}s do {B}rasil e sua marca{\c{c}}{\~a}o ret{\'o}rica.
\newblock Technical report, Technical Report.

\bibitem[\protect\citename{Pardo and Nunes}2004]{pardo:relaccoes:2004}
Thiago A.~S. Pardo and Maria das Gra{\c{c}}as~Volpe Nunes.
\newblock 2004.
\newblock Rela{\c{c}}{\~o}es ret{\'o}ricas e seus marcadores superficiais:
  An{\'a}lise de um corpus de textos cient{\'\i}ficos em {P}ortugu{\^e}s do
  {B}rasil.
\newblock {\em Relat{\'o}rio T{\'e}cnico NILC}.

\bibitem[\protect\citename{Pardo and Nunes}2008]{pardo:development:2008}
Thiago A.~S. Pardo and Maria das Gra{\c{c}}as~Volpe Nunes.
\newblock 2008.
\newblock On the development and evaluation of a {B}razilian {P}ortuguese
  discourse parser.
\newblock {\em Revista de Inform{\'a}tica Te{\'o}rica e Aplicada},
  15(2):43--64.

\bibitem[\protect\citename{Pardo and Seno}2005]{pardo:rhetalho:2005}
Thiago A.~S. Pardo and Eloize R.~M. Seno.
\newblock 2005.
\newblock Rhetalho: Um corpus de referência anotado retoricamente.
\newblock In {\em Proceedings of Encontro de Corpora}.

\bibitem[\protect\citename{Pitler \bgroup et al.\egroup
  }2009]{pitler:automati:2009}
Emily Pitler, Annie Louis, and Ani Nenkova.
\newblock 2009.
\newblock Automatic sense prediction for implicit discourse relations in text.
\newblock In {\em Proceedings of ACL-IJCNLP}.

\bibitem[\protect\citename{Polyak and Juditsky}1992]{polyak-1992}
Boris~T. Polyak and Anatoli~B. Juditsky.
\newblock 1992.
\newblock Acceleration of stochastic approximation by averaging.
\newblock {\em SIAM J. Control Optim.}, 30(4):838--855, July.

\bibitem[\protect\citename{Redeker \bgroup et al.\egroup
  }2012]{redeker:multi:2012}
Gisela Redeker, Ildikó Berzlánovich, Nynke van~der Vliet, Gosse Bouma, and
  Markus Egg.
\newblock 2012.
\newblock Multi-layer discourse annotation of a dutch text corpus.
\newblock In {\em Proceedings of LREC}.

\bibitem[\protect\citename{Roze}2013]{roze:algebre:2013}
Charlotte Roze.
\newblock 2013.
\newblock {\em Vers une alg{\`e}bre des relations de discours}.
\newblock {Ph.D.} thesis, Universit{\'e} Paris-Diderot.

\bibitem[\protect\citename{Rutherford and
  Xue}2014]{rutherford:discovering:2014}
Attapol Rutherford and Nianwen Xue.
\newblock 2014.
\newblock Discovering implicit discourse relations through brown cluster pair
  representation and coreference patterns.
\newblock In {\em Proceedings of EACL}.

\bibitem[\protect\citename{Sagae and Lavie}2005]{sagae-2005}
Kenji Sagae and Alon Lavie.
\newblock 2005.
\newblock A classifier-based parser with linear run-time complexity.
\newblock In {\em Proceedings of the Ninth International Workshop on Parsing
  Technology}, pages 125--132. Association for Computational Linguistics.

\bibitem[\protect\citename{Sagae}2009]{sagae:analysis:2009}
Kenji Sagae.
\newblock 2009.
\newblock Analysis of discourse structure with syntactic dependencies and
  data-driven shift-reduce parsing.
\newblock In {\em Proceedings of IWPT 2009}.

\bibitem[\protect\citename{Soricut and Marcu}2003]{soricut:sentence:2003}
Radu Soricut and Daniel Marcu.
\newblock 2003.
\newblock Sentence level discourse parsing using syntactic and lexical
  information.
\newblock In {\em Proceedings of NAACL}.

\bibitem[\protect\citename{Sporleder and Lapata}2005]{sporleder:discourse:2005}
Caroline Sporleder and Mirella Lapata.
\newblock 2005.
\newblock Discourse chunking and its application to sentence compression.
\newblock In {\em Proceedings of HLT/EMNLP}.

\bibitem[\protect\citename{Stede and Neumann}2014]{stede:postdam:2014}
Manfred Stede and Arne Neumann.
\newblock 2014.
\newblock Potsdam commentary corpus 2.0: Annotation for discourse research.
\newblock In {\em Proceedings of LREC}.

\bibitem[\protect\citename{Stede}2004]{stede:postdam:2004}
Manfred Stede.
\newblock 2004.
\newblock The potsdam commentary corpus.
\newblock In {\em Proceedings of the ACL Workshop on Discourse Annotation}.

\bibitem[\protect\citename{Subalalitha and
  Parthasarathi}2012]{subalalitha:approach:2012}
C~N Subalalitha and Ranjani Parthasarathi.
\newblock 2012.
\newblock An approach to discourse parsing using sangati and {R}hetorical
  {S}tructure {T}heory.
\newblock In {\em Proceedings of the Workshop on Machine Translation and
  Parsing in Indian Languages (MTPIL-2012)}.

\bibitem[\protect\citename{Subba and Di~Eugenio}2009]{subba:effective:2009}
Rajen Subba and Barbara Di~Eugenio.
\newblock 2009.
\newblock An effective discourse parser that uses rich linguistic information.
\newblock In {\em Proceedings of ACL-HLT}.

\bibitem[\protect\citename{Taboada and Mann}2006]{taboada:applications:2006}
Maite Taboada and William~C. Mann.
\newblock 2006.
\newblock Applications of rhetorical structure theory.
\newblock {\em Discourse Studies}, 8:567--588.

\bibitem[\protect\citename{Thione \bgroup et al.\egroup
  }2004]{thione:hybrid:2004}
Gian~Lorenzo Thione, Martin Van~den Berg, Livia Polanyi, and Chris Culy.
\newblock 2004.
\newblock Hybrid text summarization: Combining external relevance measures with
  structural analysis.
\newblock In {\em Proceedings of the ACL Workshop Text Summarization Branches
  Out}.

\bibitem[\protect\citename{Vliet \bgroup et al.\egroup
  }2011]{vliet:building:2011}
Nynke Van~Der Vliet, Ildik{\'o} Berzlánovich, Gosse Bouma, Markus Egg, and
  Gisela Redeker.
\newblock 2011.
\newblock Building a discourse-annotated {D}utch text corpus.
\newblock In {\em S. Dipper and H. Zinsmeister (Eds.), Beyond Semantics,
  Bochumer Linguistische Arbeitsberichte 3}, pages 157--171.

\bibitem[\protect\citename{Wu \bgroup et al.\egroup }2016]{wu:new:2016}
Yunfang Wu, Fuqiang Wan, Yifeng Xu, and Xueqiang L{\"u}.
\newblock 2016.
\newblock A new ranking method for {C}hinese discourse tree building.
\newblock {\em Acta Scientiarum Naturalium Universitatis Pekinensis},
  52(1):65--74.

\end{thebibliography}
\bibliographystyle{eacl2017}

\appendix

\section{Mapping of the relations}
\label{sec:supplemental}

\begin{table*}
\begin{tabular}{ll}
\toprule
Classe & Relations \\
\midrule
\crel{Attribution}	&	\rel{attribution}, \rel{attribution-negative}	\\
\crel{Background}	&	\rel{background}, \rel{circumstance}, \rel{circunstancia}, \rel{fondo}, \\
& \rel{preparaci{\'o}n}, \rel{preparation}, \rel{prestatzea}, \rel{testuingurua},\\
&  \rel{zirkunstantzia}	\\
\crel{Cause}	&	\rel{causa}, \rel{cause}, \rel{cause-result}, \rel{consequence}, \rel{kausa},  \\
& \rel{non-volitional-cause}, \rel{non-volitional-result}, \rel{ondorioa}, \\
&  \rel{result}, \rel{resultado}, \rel{volitional-cause}, \rel{volitional-result}	\\
\crel{Comparison}	&	\rel{analogy}, \rel{comparison}, \rel{preference}, \rel{proportion}	\\
\crel{Condition}	&	\rel{alderantzizko-baldintza}, \rel{alternativa}, \rel{aukera}, \rel{baldintza},  \\
& \rel{condici{\'o}n}, \rel{condici{\'o}n-inversa}, \rel{condition}, \rel{contingency}, \\
&  \rel{ez-baldintzatzailea}, \rel{hypothetical}, \rel{otherwise}, \\
& \rel{unconditional},  \rel{unless}	\\
\crel{Contrast}	&	\rel{antitesia}, \rel{antithesis}, \rel{ant{\'i}tesis}, \rel{concesi{\'o}n}, \rel{concession}, \\
& \rel{contrast}, \rel{contraste}, \rel{kontrastea}, \rel{kontzesioa}	\\
\crel{Elaboration}	&	\rel{definition}, \rel{e-elaboration}, \rel{elaboraci{\'o}n}, \rel{elaboration}, \\
& \rel{elaboration-additional}, \rel{elaboration-general-specific},  \\
& \rel{elaboration-object-attribute}, \rel{elaboration-part-whole},  \\
& \rel{elaboration-process-step}, \rel{elaboration-set-member},  \\
& \rel{elaborazioa}, \rel{example}, \rel{parenthetical}	\\
\crel{Enablement}	&	\rel{ahalbideratzea}, \rel{capacitaci{\'o}n}, \rel{enablement}, \rel{helburua}, \\
& \rel{prop{\'o}sito}, \rel{purpose}	\\
\crel{Evaluation}	&	\rel{comment}, \rel{conclusion}, \rel{ebaluazioa}, \rel{evaluaci{\'o}n}, \rel{evaluation}, \\
& \rel{interpretaci{\'o}n}, \rel{interpretation}, \rel{interpretazioa}	\\
\crel{Explanation}	&	\rel{ebidentzia}, \rel{evidence}, \rel{evidencia}, \rel{explanation},  \\
& \rel{explanation-argumentative}, \rel{justificaci{\'o}n}, \rel{justifikazioa}, \\
&  \rel{justify}, \rel{motibazioa}, \rel{motivaci{\'o}n}, \rel{motivation}, \rel{reason}	\\
\crel{Joint}	&	\rel{bateratzea}, \rel{conjunci{\'o}n}, \rel{conjunction}, \rel{disjunction},\\
&  \rel{disjuntzioa}, \rel{disyunci{\'o}n}, \rel{joint}, \rel{konjuntzioa}, \rel{list}, \rel{lista}, \rel{uni{\'o}n}	\\
\crel{Manner-Means}	&	\rel{manner}, \rel{means}, \rel{medio}, \rel{metodoa}	\\
\crel{Same-unit}	&	\rel{same-unit}	\\
\crel{Summary}	&	\rel{birformulazioa}, \rel{definitu-gabeko-erlazioa}, \rel{laburpena},  \\
& \rel{reformulaci{\'o}n}, \rel{restatement}, \rel{resumen}, \rel{summary}	\\
\crel{Temporal}	&	\rel{inverted-sequence}, \rel{secuencia}, \rel{sekuentzia}, \rel{sequence}, \\
& \rel{temporal-after}, \rel{temporal-before}, \rel{temporal-same-time}	\\
\crel{Textual-organization}	&	\rel{textual-organization}	\\
\crel{Topic-Change}	&	\rel{topic-drift}, \rel{topic-shift}	\\
\crel{Topic-Comment}	&	\rel{arazo-soluzioa}, \rel{comment-topic}, \rel{problem-solution},  \\
& \rel{question-answer}, \rel{rhetorical-question}, \rel{soluci{\'o}n}, \\
&  \rel{solutionhood}, \rel{statement-response}, \rel{topic-comment}	\\
\bottomrule
\end{tabular}
\caption{Mapping of all the relations found in the datasets: for each class, we give the set of relation names as they appear in the corpora (removing only the possible suffixes ``-e", ``-s", ``-mn"). We ignore the simplest differences in names (e.g. \rel{textual-organization} and \rel{textualorganization}).}
\end{table*}

\end{document}